\documentclass{article}

\usepackage{arxiv}

\usepackage[utf8]{inputenc} 
\usepackage[T1]{fontenc}    
\usepackage{hyperref}       
\usepackage{url}            
\usepackage{booktabs}       
\usepackage{amsfonts}       
\usepackage{nicefrac}       
\usepackage{microtype}      
\usepackage{lipsum}		
\usepackage{graphicx}
\usepackage{doi}
\usepackage{amsmath,amssymb,amsfonts}
\usepackage[square,numbers]{natbib}
\usepackage{amsthm}%
\usepackage{mathrsfs}%
\usepackage[title]{appendix}%
\usepackage{xcolor}%
\usepackage{textcomp}%
\usepackage{manyfoot}%
\usepackage{booktabs}%
\usepackage{hyperref}
\usepackage{url}
\usepackage{subcaption}

\usepackage[ruled,vlined,linesnumbered]{algorithm2e}
\usepackage{listings}%
\usepackage{booktabs}

\title{MiNER: Fine-Tuned Biomedical Natural Language Processing for Malaria Disease Entity Recognition in Clinical Texts}

\date{} 					

\author{ \href{https://orcid.org/0000-0001-6673-6932}{\includegraphics[scale=0.06]{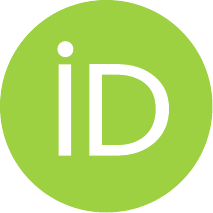}\hspace{1mm}V. S. Anoop} \\
	Department of Computer Science and Engineering, Amrita School of Computing\\
	Amrita Vishwa Vidyapeetham\\
	Kollam, India \\
	\texttt{anoopvs@am.amrita.edu} \\
	\And
	\href{https://orcid.org/0000-0000-0000-0000}{\includegraphics[scale=0.06]{orcid.pdf}\hspace{1mm}Devika N.} \\
	School of Digital Sciences\\
	Kerala University of Digital Sciences, Innovation and Technology\\
	Thiruvananthapuram, India \\
	\texttt{devika.ds21@duk.ac.in } \\
}

\renewcommand{\shorttitle}{MiNER: Biomedical NLP for Malaria Disease Entity Recognition}

\hypersetup{
pdftitle={MiNER: Biomedical NLP for Malaria Disease Entity Recognition},
pdfauthor={V. S. Anoop, Devika N.},
pdfkeywords={Natural language processing, biomedical, language models, biomedical information extraction, biomedical named entity extraction, malaria disease},
}

\begin{document}
\maketitle

\begin{abstract}
	Malaria remains a significant global health burden, necessitating continuous research efforts to understand its complex molecular mechanisms, epidemiology, and potential therapeutic interventions. Extracting essential biomedical information from the vast and constantly growing malaria literature is a challenging task that demands innovative approaches. Recently, pre-trained language models have revolutionized natural language processing tasks, demonstrating remarkable capabilities in various domains. This paper proposes a fine-tuned pre-trained biomedical language model for biomedical information extraction from scientific literature on malaria disease. The proposed methodology selects and preprocesses a large corpus of scientific articles on malaria, and then annotates them with entities of clinical significance. It then leverages BioBERT, a state-of-the-art pre-trained language model, to encode the textual data into context-aware representations. We fine-tune the model using domain-specific annotations and supervised learning to enhance its ability to extract relevant biomedical named entities. Extensive experiments and comparisons with different encoding and machine learning algorithms show that the proposed approach significantly outperforms them in precision, recall, and accuracy. We also publish our human-labeled dataset for entity and relation extraction to enable other health informatics researchers to train advanced models for malaria information extraction.
\end{abstract}


\section{Introduction}
Natural Language Processing (NLP), a subfield of Artificial Intelligence (AI), allows computer systems to comprehend natural language (both text and speech) as we humans do and represent and analyze human language computationally\cite{khurana2023natural}. NLP helps computers understand and process complex human language, which is often dynamic and complex. It has a wide range of applications and use-cases such as sentiment analysis\cite{anoop2024graph}, machine translation\cite{leiter2024towards}, information extraction\cite{lim2024test}, text classification\cite{chandran2023topicstriker}, named entity recognition\cite{hu2024improving}\cite{rajimol2020framework}, question answering systems\cite{anoop2016learning}, chatbots and virtual assistants\cite{kurniawan2024systematic}, text summarization \cite{van2024adapted}, and language generation\cite{lovelace2024latent}. Information extraction involves extracting meaningful information from large amounts of unstructured text extensively spread across various resources on the web and other platforms. This area of natural language processing has gained popularity among researchers as the need for effective information extraction has become increasingly important. Earlier approaches to information extraction relied on rule-based systems and statistical models; however, this field of NLP has undergone a significant transformation in recent years with the emergence of pre-trained language models \ cite {son2024ftmmr}, \cite{wang2024hct}, and \cite{anoop2019extracting}. These text-understanding models with contextual understanding capabilities transformed the way in which machines understand and comprehend human languages\cite{wu2025survey}\cite{gautam2025survey}\cite{kalyan2022ammu}.\\ 

Malaria, one of the deadliest diseases caused by the Plasmodium parasite and transmitted through the bite of infected mosquitoes, continues to pose a formidable threat to global health\cite{odero2024early}\cite{verma2024deep}\cite{marsh2024development}. In the recent past, the medical fraternity has made significant progress in combating malaria through different interventions at a global scale. But still, the complexity of the disease demands significant efforts to synthesize knowledge to help the researchers understand and develop effective strategies for prevention, diagnosis, and treatment. The wealth of scientific knowledge that is latent within the vast corpus of malaria literature has huge potential to help healthcare researchers. However, the huge volume of research papers, clinical trials, case studies, and epidemiological reports presents a significant challenge in extracting valuable information\cite{lopez2025clinical}\cite{gu2025scalable}. Manual collection, curation, and analysis of such unstructured data is a laborious and time-consuming process and is prone to errors. Recent advances in natural language processing, particularly the development of pre-trained language models, have demonstrated remarkable performance in various tasks by leveraging extensive knowledge from vast datasets. Models such as BERT\cite{devlin2018bert}, GPT\cite{floridi2020gpt}, and RoBERTa\cite{liu2019roberta} have revolutionized natural language processing and have been applied to various domains such as biomedical research\cite{devika2023biomedical}\cite{ardra2023oralmedner}.\\

This paper explores the potential of using domain-specific pre-trained language models for biomedical information extraction from the vast collection of scientific literature on malaria. We aim to accelerate the pace of malaria research and facilitate evidence-based decision-making, which may ultimately contribute to global efforts in combating this disease. The proposed methodology collects a large amount of unstructured data on Malaria from biomedical literature repositories such as PubMed and employs a manual effort for labeling the entities. The domain-specific biomedical pre-trained model BioBERT is then used to obtain the contextual embedding to train different machine learning algorithms. Furthermore, the results of our experiments and evaluations will be presented, providing insights into the effectiveness of pre-trained language models in identifying and extracting key entities from a large collection of unstructured text. This paper will address potential future directions and opportunities to enhance the performance and applicability of our proposed information extraction approach and publish the labeled dataset for other health informatics researchers. The major contributions of this paper may be summarized as follows:
\begin{itemize}
    \item Proposes an approach using a domain-specific biomedical language model for extracting clinical entities on Malaria disease.
    \item Experimentally verifies the proposed approach using systematic experiments and compares the results with state-of-the-art machine learning approaches.
    \item Publish the labeled dataset and code for other natural language processing researchers and practitioners to train advanced NER models.
\end{itemize}
\indent The subsequent sections are organized as follows: Section 2 discusses the related works, and in Section 3, the materials and methods are detailed. In Section 4, the proposed approach for biomedical named entity recognition is detailed. Section 5 presents the results and discussions, and finally, the conclusions and future directions are discussed in Section 6.
\section{Related Studies}
This section discusses some of the prominent and recent approaches reported in the natural language processing literature that use pre-trained language models to extract biomedical information from textual data. Recent advances in natural language processing, such as deep learning and large language models, have led to the development of better context-understanding models. The recognition of named entities in the biomedical domain is considered one of the most challenging tasks due to many other factors, such as the limited availability of freely accessible data, the annotation of data to train models, and the extensive coverage of biomedical concepts. In recent years, there have been some prominent works reported with varying degrees of accuracy that attempt to address these challenges\cite{nair2024hey}.\\

\indent Biomedical named entity recognition approaches heavily rely on machine learning methods, a conventional approach where the process of feature engineering can be quite time-consuming\cite{goyal2025named}. The goal of biomedical named entity recognition is to accurately assign labels to words within input sequences, a critical task requiring sufficient labeled data, as it is a supervised sequence classification problem. Addressing the challenge of capturing multi-word entities necessitates the use of models that can represent sequences with tags like Beginning (B), Inside (I), Outside (O), and Ending (E), adhering to \textit{IOB}, \textit{IOE}, or \textit{IOBES} tagging standards. For example, a single-word entity resembling a disease is labeled as "S-Disease," whereas multi-word entities are tagged as "B-Disease," "I-Disease," and "E-Disease," denoting the beginning, inside, and end segments of the entity, respectively\cite{yang2025large}.\\

\indent Recent advancements in biomedical named entity extraction have witnessed the adoption of deep learning techniques, such as convolutional neural networks, long short-term memory networks, and the state-of-the-art transformer-based language models \cite{sreenivas2025enhancing}\cite{lopez2025clinical}. There are also several methods published that use hybrid approaches integrating traditional deep neural networks with attention mechanisms\cite{parsaeimehr2023improving}. While the deep neural networks capture specific word features and build contextual understanding within input sentences\cite{parsaeimehr2023improving}. Some architectures combine a pre-trained model with long short-term memory networks through transfer learning that outperformed baseline models\cite{usha2022named}. This model incorporates a Bi-LSTM framework alongside a conditional random field (CRF) layer, effectively modeling the dependencies among states within entire input sequences. Notably, it achieved impressive performance metrics with a 91\% f1-score and 98\% accuracy\cite{usha2022named}.\\

\indent The traditional named entity recognition models rely heavily on labor-intensive feature engineering. The pretrained embeddings, on the other hand, use semantic context-aware feature embedding at both character level and word level, which enhances the recognition accuracy  \cite{siddalingappa2022bi}. The field has seen further advancements with the adoption of deep learning techniques and statistical embeddings, such as LSTM-CRF, resulting in substantial improvements in identifying biomedical entities\cite{lee2020biobert}. While multitask strategies have shown effectiveness in disease entity recognition, recent studies suggest augmenting these approaches with transfer learning using models like BERT combined with CRF for superior performance\cite{jarashanth2022applying}. Moreover, transformer-based models like BERT and domain-specific BioBERT have been integrated into the embedding layer, surpassing previous BiLSTM+CRF architectures in accuracy and effectiveness. These advancements underscore the evolving landscape of biomedical named entity recognition that leverages deep learning and contextual embeddings to achieve new benchmarks in biomedical text analysis.\\ 

\indent There are many recent studies that have focused on enhancing named entity recognition models tailored for biomedical data. Some studies have highlighted the potential of transfer learning with BERT, incorporating Conditional Random Fields (CRF), resulting in superior performance compared to traditional deep learning and multitask learning approaches\cite{jarashanth2022applying}. Some studies demonstrate that deep learning techniques such as Recurrent Neural Networks (RNNs) and Long Short-Term Memory Networks (LSTMs) achieve better results than baseline machine learning models\cite{siddalingappa2022bi}. However, these methods initially struggled with capturing relevant information from lengthy text sequences due to the long-range dependency problem. To overcome these challenges, leveraging pretrained models is crucial. These models, trained on specific domains, excel in capturing contextual information pertinent to their training domain \cite{kalyan2022ammu}. Despite the availability of general domain models, pretrained models that are specifically tailored for biomedical texts are limited. BioBERT, a domain-specific language model pretrained on a vast biomedical corpus, has demonstrated significant improvements in various text mining tasks, including biomedical NER, question answering, and relation extraction \cite{lee2020biobert}. The proposed approach uses the capabilities of BioBERT, which is a biomedical pretrained model and finetunes it for the identification of malaria-related named entities.\\
\begin{figure*}[ht!]
\centering
\includegraphics[width=\textwidth]{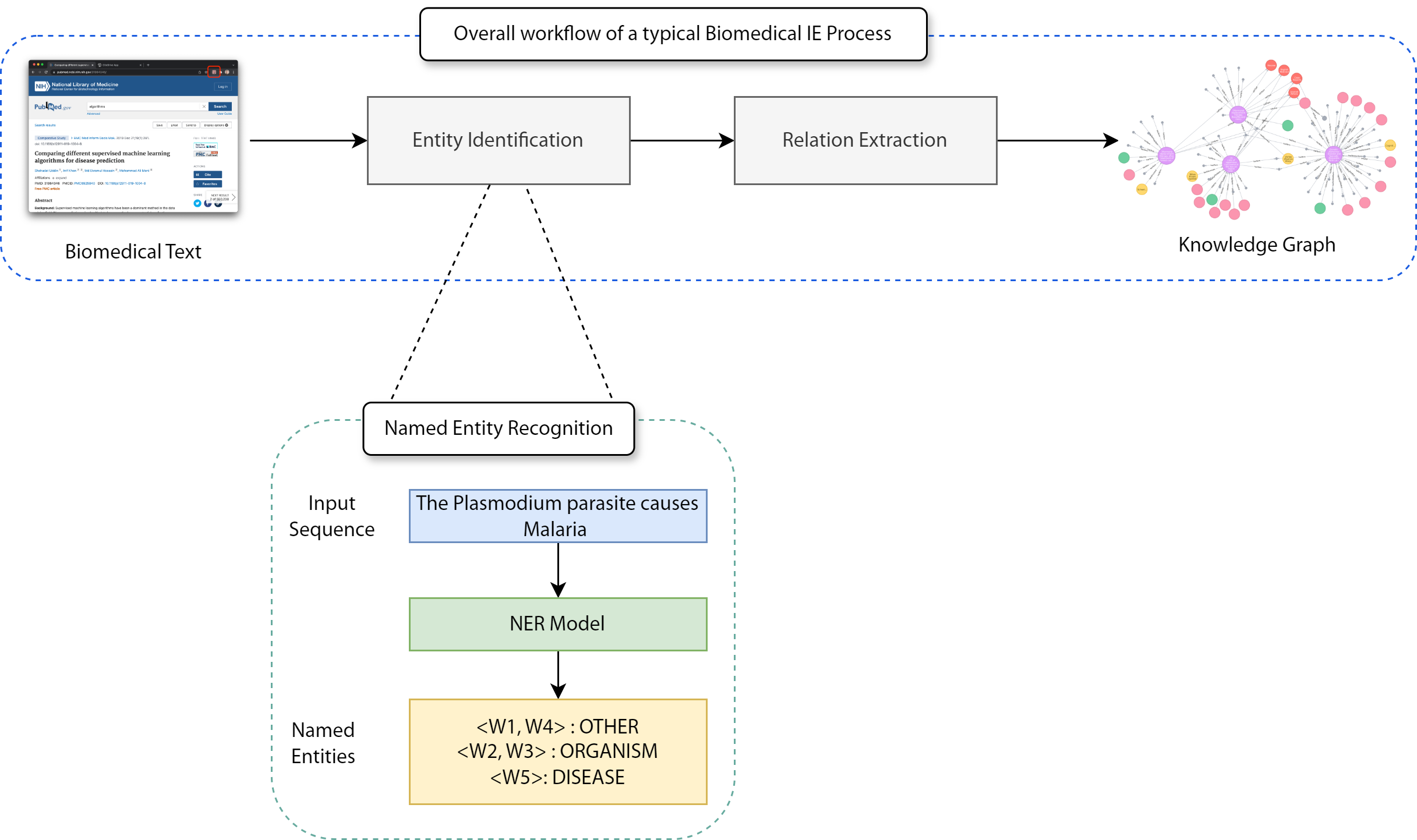}
\caption{An illustration of the proposed Named Entity Recognition model}
\end{figure*}
\section{Materials and Methods}
\indent Named entity recognition in the Biomedical domain is a crucial task and faces many challenges. Early approaches towards NER use various baseline shallow machine-learning models to classify the named entities. But sometimes, these models fail to address complex and multi-word entities. Deep learning approaches require minimal feature engineering and hold great promise. Many datasets are available for this task in the generic domain, but domain-specific datasets are few. So, for a domain-specific task, labeled data is very limited. Even though many sources like PubMed and various other databases have a lot of information, extracting and manually annotating them is difficult, as they carry vast concepts, and annotation needs expert knowledge of biomedical concepts. This section lists and discusses some materials and methods used to implement the proposed approach.
\subsection{Label Studio}
Label Studio (available at \url{https://labelstud.io/}) is a free data labeling tool that assists in annotating and labeling various types of data, such as text, images, audio, and video. It offers an intuitive interface for generating and organizing labeling assignments, enabling users to establish labeling structures, invite contributors, and monitor the advancement of labeling endeavors. Data labeling is a crucial and challenging step in machine learning. It is time-consuming, prone to human errors, and requires managing multiple annotators. Label Studio addresses these challenges by providing a user-friendly interface, supporting various annotation types, enabling collaboration, and offering quality control features. It streamlines the process, enhances annotation quality, and contributes to the development of robust models.
\subsection{Feature Encoding}
The labeled and preprocessed data is then encoded to numerical vectors using various feature encoding techniques as follows. These features are trained using various machine learning classifiers to classify and identify the named entities.
\begin{itemize} 
\item \textbf{TF-IDF}: Term Frequency-Inverse Document Frequency is a widely used numerical statistic in text mining and information retrieval. It measures the importance of a term in a document by considering its frequency in the document and its rarity in the corpus. The TF-IDF score, $TF-IDF(t, d, D)$ is calculated as the product of the term frequency $TF(t, d)$ and the inverse document frequency $IDF(t, D)$. $TF$ represents how often the term appears in the document, while $IDF$ measures the uniqueness of the term across the corpus. The formula for computing the $TF-IDF$ score is shown in Eq (1).
\begin{equation}
\text{TF-IDF}(t, d, D) = \text{TF}(t, d) \times \text{IDF}(t, D)
\end{equation}
\item \textbf{Count Vectors}: also known as Bag-of-Words (BoW) representation, are vectors representing text documents as a count of each word occurrence. Each document is represented by a vector, with each element representing the number of times a certain word appears in the document. Count vectors disregard word order and context, instead capturing the presence or absence of words that are specified in the document.
\item \textbf{GloVe}: It is a word embedding learning system that learns from a vast corpus of text. Word embeddings are compact vector representations that capture the semantic relationships of words. GloVe generates word vectors from global word co-occurrence statistics, which can then be used to encode words in a text document.
\item \textbf{Word2Vec}: It is another popular unsupervised learning algorithm for word embeddings that learns distributed representations of words based on their contextual usage. Word2Vec generates word vectors by predicting the probability of a word given its context or predicting the context given the word. These word vectors can be used to encode words in a text document.
\item \textbf{BERT}: is a transformer-based model that learns word representations by taking into account the context from both preceding and following words. It is trained on large amounts of textual data through masked language modeling and next-sentence prediction tasks. BERT effectively captures the contextual information of words and can be further adapted for various NLP tasks.
\item \textbf{BioBERT}: It is a specialized version of BERT specifically designed for biomedical text data. It undergoes pre-training using a methodology similar to BERT but on a corpus of biomedical text, making it particularly well-suited for biomedical natural language processing tasks. BioBERT offers the flexibility to fine-tune its pre-trained model for various biomedical tasks, such as identifying named entities, extracting relationships, and answering biomedical questions, and is used in the proposed approach.
\subsection{Machine Learning Algorithms}
\begin{itemize}
\item \textbf{Support Vector Machine (SVM)}: is a commonly applied and highly efficient machine learning algorithm utilized for both classification and regression tasks. It finds an optimal hyperplane to separate different classes with maximum margin. SVM has the ability to handle both linear and non-linear classification problems through the utilization of diverse kernel functions.
\item \textbf{Logistic Regression (LR)}: is a widely utilized statistical model extensively employed for binary classification tasks. Although its name suggests regression, it is primarily utilized for classification purposes. Logistic Regression demonstrates strong performance when the decision boundary is linear or can be reasonably approximated by a linear function.
\item \textbf{Random Forest (RF)}: is a popular ensemble learning technique that utilizes multiple decision trees to make predictions. This technique trains each tree within the forest on a randomly selected subset of data and features. This randomization helps mitigate the overfitting risk, enhancing the model's overall performance and generalization ability.
\item \textbf{Naive Bayes (NB)}: is a straightforward and probabilistic classification algorithm that relies on Bayes' theorem. It operates under the assumption of conditional independence among features given the class label, hence the term "naive." Through the computation of probabilities for each class based on the input features, this algorithm identifies the class with the highest probability and assigns it as the predicted outcome.
\subsection{Evaluation Matrices}
The performance evaluation of the proposed named entity classification approach was done using standard performance metrics such as accuracy, precision, recall, and F1-score. The formulas used to calculate these metrics are provided in Eq. (2), Eq. (3), Eq. (4), and Eq. (5).
\begin{equation}
\textbf{Accuracy} = \frac{\text{TP} + \text{TN}}{\text{TP} + \text{TN} + \text{FP} + \text{FN}}
\end{equation}
\begin{equation}
\textbf{Precision} = \frac{\text{TP}}{\text{TP} + \text{FP}}
\end{equation}
\begin{equation}
\textbf{Recall} = \frac{\text{TP}}{\text{TP} + \text{FN}}
\end{equation}
\begin{equation}
\textbf{F-measure} = \frac{2 \times \text{Precision} \times \text{Recall}}{\text{Precision} + \text{Recall}}
\end{equation}
\end{itemize}
\end{itemize} 
\section{Proposed Approach}
This section discusses the proposed approach for the named entity recognition of malaria disease. The overall workflow of a typical biomedical information extraction pipeline, with a specific focus on our named entity recognition, is given in Fig. 1. The dataset used for this experiment is taken from PubMed\footnote{https://pubmed.ncbi.nlm.nih.gov/}, which is an online database that provides access to millions of medical literature. The PubMed abstracts on Malaria were collected using PyMed library, which is a Python library that provides access to PubMed through the PubMed API, and the collected abstracts were manually annotated using \textbf{Label Studio}, a data labeling and annotation tool that is used for creating, managing, and labeling data for machine learning tasks. For generic NER tasks, labels like name, place, and organization are used. For our domain-specific NER task for Malaria texts, we need specific labels particular to our domain. The chosen labels were Disease, Organism, Medication, Protein, Gene, Anatomical Structures, Chemical Structures, and Other(O). As NER encounters a multiword capturing problem, it cannot correctly classify the multiword entities. To achieve this, the data is labeled using IOBES tagging standards, where "B", "I", and "E" represents the "Beginning", "Inside", and "End" of an entity, "S" represents a "single word" entity and all other words irrelevant to the domain are labeled as "O". The annotated data is further preprocessed and used for training various models. Fig. 3 shows the input text and the output from our proposed approach with labeled entities.\\
\begin{algorithm}[t]
\DontPrintSemicolon
\caption{Algorithm for the proposed named entity recognition approach}
\label{alg:ner}
\KwIn{Labeled corpus $\mathcal{D}=\{(x^{(i)}, y^{(i)})\}_{i=1}^N$ with BIO tags.}
\KwOut{Best embedding-classifier pair $(E^*, C^*)$ with evaluation metrics.}

\textbf{Step 1: Preprocessing} \;
- Normalize and tokenize text. \;
- Align tokens with BIO labels. \;
- Split data into train/dev/test. \;

\textbf{Step 2: Feature Extraction} \;
Define embedding and feature set $\mathcal{E} = \{$TF-IDF, CountVector, Word2Vec, GloVe, BERT, BioBERT$\}$. \;
\For{$E \in \mathcal{E}$}{
  Transform $x^{(i)}$ into feature vectors $f^{(i)} = E(x^{(i)})$. \;
}

\textbf{Step 3: Model Training} \;
Define classifier set $\mathcal{C} = \{$SVM, LR, RF, and NB$\}$. \;
\For{$E \in \mathcal{E}$}{
  \For{$C \in \mathcal{C}$}{
    Train classifier $C$ on features $\{f^{(i)}\}$ with labels $\{y^{(i)}\}$. \;
    Perform $k$-fold cross-validation. \;
    Compute entity-level Precision, Recall, and F1-score. \;
    Store results as $(E, C, \mathrm{F1})$. \;
  }
}
\textbf{Step 4: Final Evaluation} \;
Retrain $(E, C)$ on train+dev set. \;
Evaluate on held-out test set. \;
Report entity-level Precision, Recall, and F1. \;
\end{algorithm}
\begin{figure}[ht!]
\centering
\includegraphics[width=8.5cm]{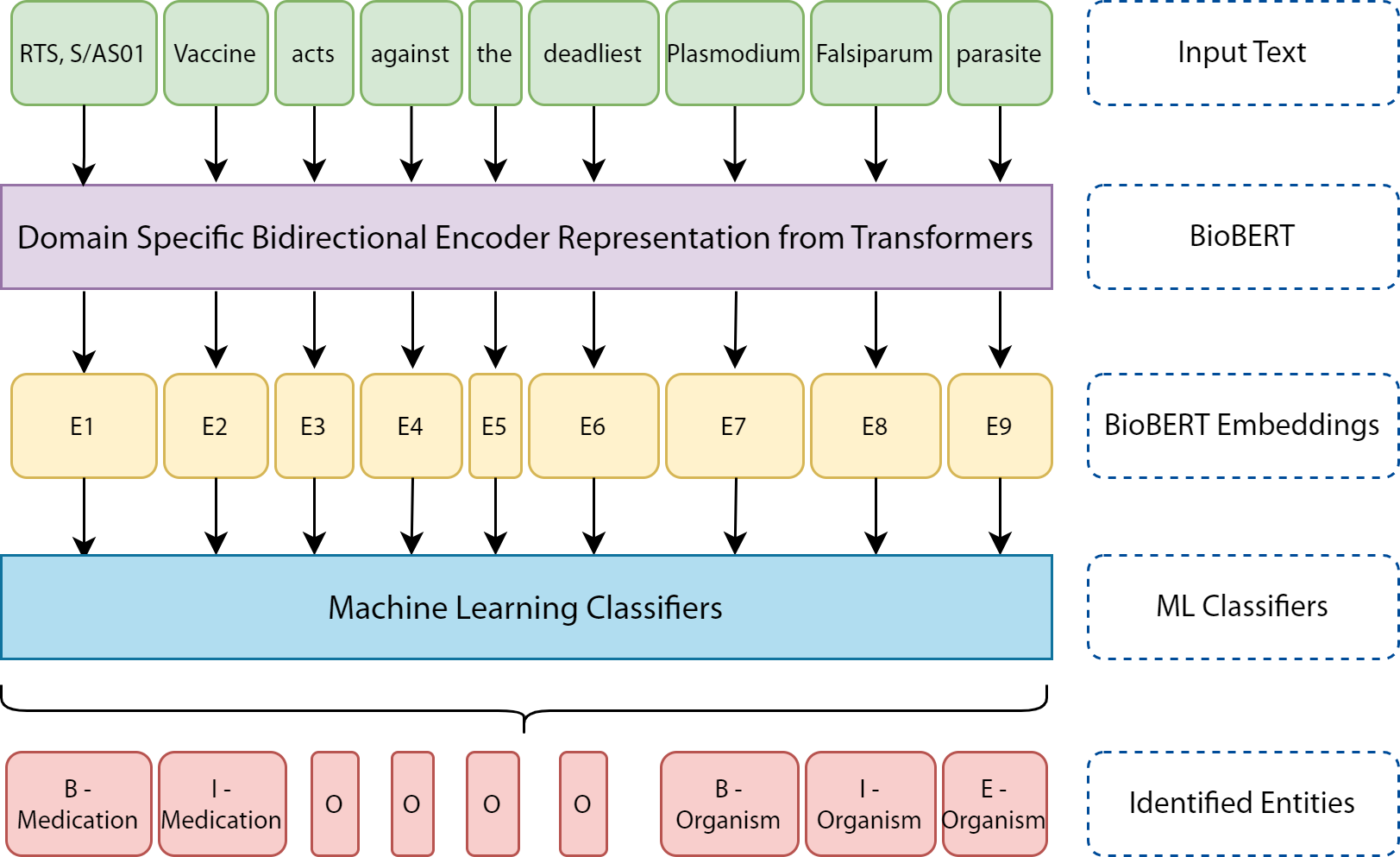}
\caption{Architecture of the proposed approach}
\end{figure}
\indent After necessary preprocessing, the dataset is converted to numerical representations using various feature encoding techniques like TF-IDF, Count Vectors, Word2vec, Glove, BERT, and BioBERT. A detailed description of these techniques is provided in the previous chapter. Once the feature encoding is complete, the resulting feature vectors are used as inputs for various machine learning models, including SVM, Random Forest, Logistic Regression, and Naive Bayes. Fig. 2 illustrates the overall architecture used in this NER system. Here, we can see that the input sequence is split into individual words, which are then converted to numerical representations using BERT and BioBERT embeddings. After this, they are fed as the input for the machine learning classifiers. This will classify the input into predefined classes. This model is deployed as an application using Streamlit, an open-source Python library used for building interactive web applications and data dashboards. The authors of this paper publish the labeled dataset and associated code files for other biomedical and natural language processing researchers to train advanced models. The published dataset may be accessed from the URL \url{https://github.com/anoop-vs/nlp-vector-borne-diseases}.
\begin{figure}[ht!]
\centering
\includegraphics[width=\textwidth, height=7cm]{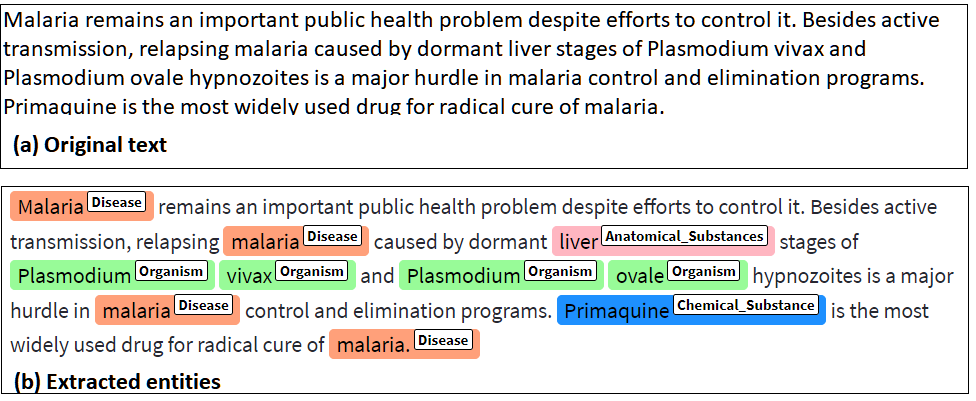}
\caption{The output from the proposed system for Malaria named entity recognition}
\end{figure}
\section{Results and Discussions}
The results obtained from the experiment, as discussed in the methodology section, are presented here. The Precision, Recall, Accuracy, and F-measure (F) values for different classifiers, such as Support Vector Machine (SVM), Logistic Regression (LR), Random Forest classifier (RF), and Naive Bayes classifier (NB) for different features are shown in Table 1, Table 2, and Table 3. Also, the graphs showing the performance comparison of different machine learning classifiers on various text feature encoding techniques, such as TF-IDF and BERT, are shown in Fig. 4, Fig. 5, Fig. 6, Fig. 7, Fig. 8, and Fig. 9.

\begin{figure*}[ht!]
    \centering
    \begin{subfigure}[t]{0.32\textwidth}
        \centering
        \includegraphics[width=\linewidth]{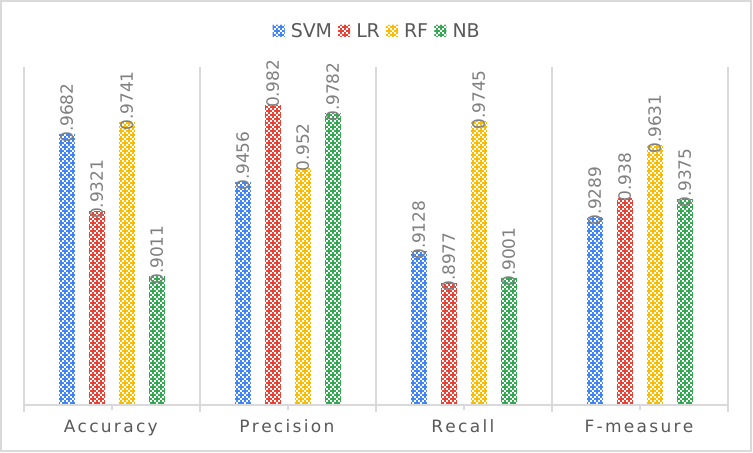}
        \caption{Precision, recall, accuracy, and f-measure for the TF-IDF feature encoding}
    \end{subfigure}
    \hfill
    \begin{subfigure}[t]{0.32\textwidth}
        \centering
        \includegraphics[width=\linewidth]{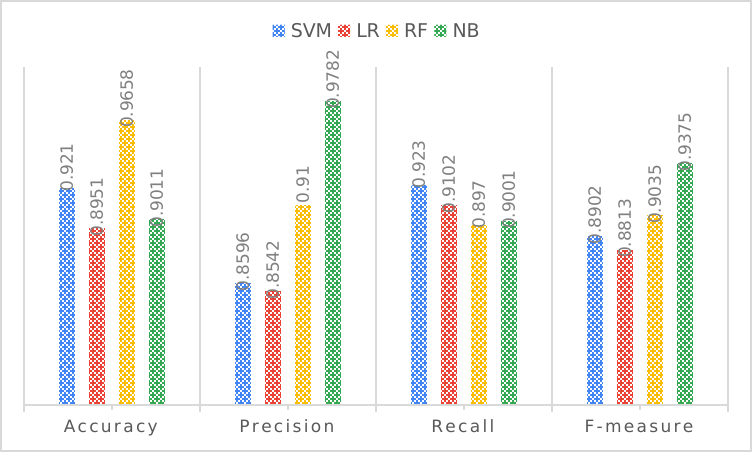}
        \caption{Precision, recall, accuracy, and f-measure for the Count Vector feature encoding}
    \end{subfigure}
    \hfill
    \begin{subfigure}[t]{0.32\textwidth}
        \centering
        \includegraphics[width=\linewidth]{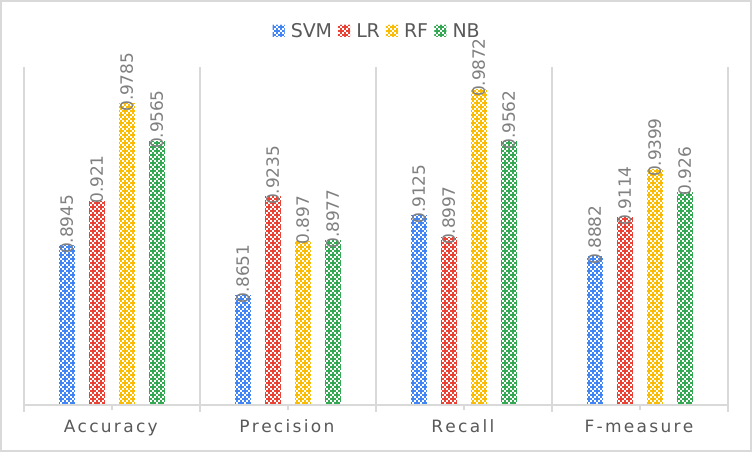}
        \caption{Precision, recall, accuracy, and f-measure for the Word2Vec feature encoding}
    \end{subfigure}
    \caption{Precision, recall, and accuracy performance of SVM, Logistic Regression, Random Forest, and Naive Bayes classifiers across different feature encoding techniques}
    \label{fig:comparison}
\end{figure*}

\begin{figure*}[ht!]
    \centering
    \begin{subfigure}[t]{0.32\textwidth}
        \centering
        \includegraphics[width=\linewidth]{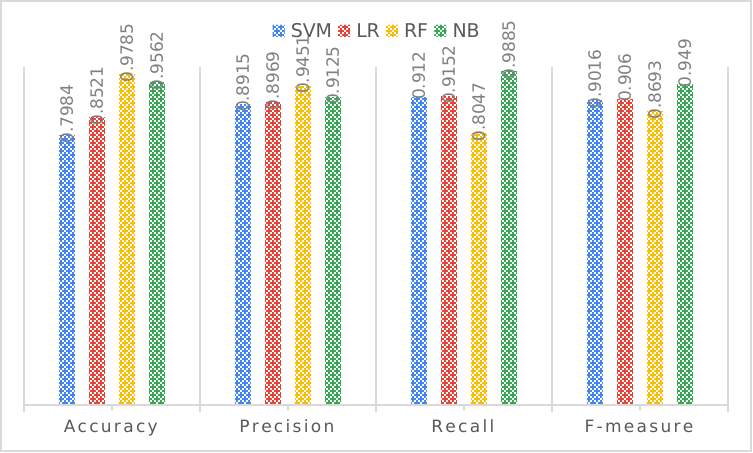}
        \caption{Precision, recall, accuracy, and f-measure for the GloVe feature encoding}
    \end{subfigure}
    \hfill
    \begin{subfigure}[t]{0.32\textwidth}
        \centering
        \includegraphics[width=\linewidth]{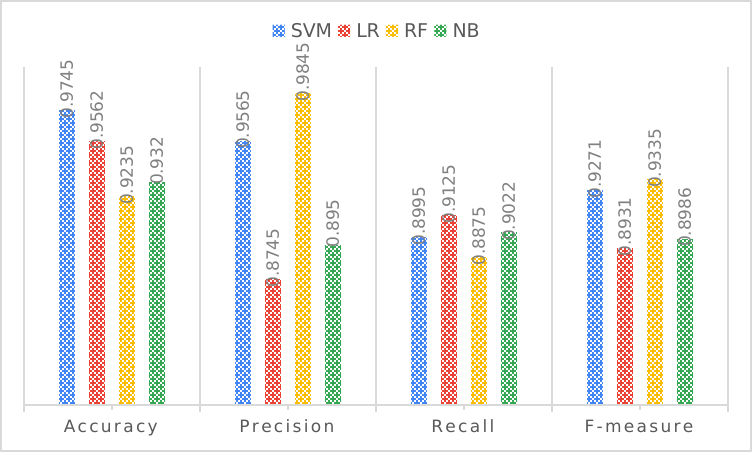}
        \caption{Precision, recall, accuracy, and f-measure for the BERT feature encoding}
    \end{subfigure}
    \hfill
    \begin{subfigure}[t]{0.32\textwidth}
        \centering
        \includegraphics[width=\linewidth]{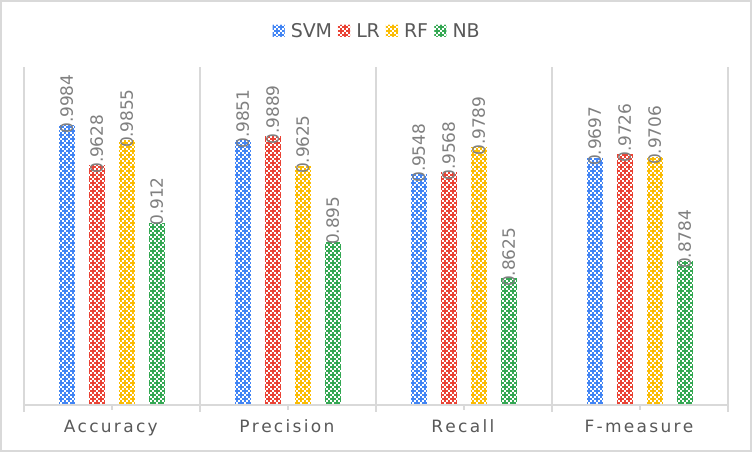}
        \caption{Precision, recall, accuracy, and f-measure for the BioBERT feature encoding}
    \end{subfigure}
    \caption{Precision, recall, and accuracy performance of SVM, Logistic Regression, Random Forest, and Naive Bayes classifiers across GloVe, BERT, and BioBERT feature encoding techniques}
    \label{fig:deep_embeddings}
\end{figure*}

\begin{table*}[h]
\caption{Precision, Recall, Accuracy, and F-measure for Support Vector Machine, Logistic Regression, Random Forest, and Naive Bayes classifiers for TF-IDF and Count Vector features}
\centering
\begin{tabular}{@{}lllllllll@{}}
\toprule
\textbf{Model} & \multicolumn{4}{c}{\textbf{TF-IDF}} & \multicolumn{4}{c}{\textbf{CV}} \\
\cmidrule(lr){2-5} \cmidrule(lr){6-9}
& \textbf{A} & \textbf{P} & \textbf{R} & \textbf{F} & \textbf{A} & \textbf{P} & \textbf{R} & \textbf{F} \\
\midrule
SVM & 0.9682 & 0.9456 & 0.9128 & 0.9289 & 0.921 & 0.8596 & 0.923 & 0.8902\\
LR & 0.9321 & 0.982 & 0.8977 & 0.938 & 0.8951 & 0.8542 & 0.9102 & 0.8813 \\
RF & 0.9741 & 0.952 & 0.9745 & 0.9631 & 0.9658 & 0.91 & 0.897 & 0.9035 \\
NB & 0.9011 & 0.9782 & 0.9001 & 0.9375 & 0.9011 & 0.9782 & 0.9001 & 0.9375 \\
\bottomrule
\end{tabular}
\end{table*}
\begin{table*}[h]
\caption{Precision, Recall, Accuracy, and F-measure for Support Vector Machine, Logistic Regression, Random Forest, and Naive Bayes classifiers for Word2Vec and Glove features}
\centering
\begin{tabular}{@{}lllllllll@{}}
\toprule
\textbf{Model} & \multicolumn{4}{c}{\textbf{Word2Vec}} & \multicolumn{4}{c}{\textbf{GloVe}} \\
\cmidrule(lr){2-5} \cmidrule(lr){6-9}
& \textbf{A} & \textbf{P} & \textbf{R} & \textbf{F} & \textbf{A} & \textbf{P} & \textbf{R} & \textbf{F} \\
\midrule
SVM & 0.8945 & 0.8651 & 0.9125 & 0.8882 & 0.7984 & 0.8915 & 0.912 & 0.9016 \\
LR & 0.921 & 0.9235 & 0.8997 & 0.9114 & 0.8521 & 0.8969 & 0.9152 & 0.906 \\
RF & 0.9785 & 0.897 & 0.9872 & 0.9399 & 0.9785 & 0.9451 & 0.8047 & 0.8693 \\
NB & 0.9565 & 0.8977 & 0.9562 & 0.926 & 0.9562 & 0.9125 & 0.9885 & 0.949 \\
\bottomrule
\end{tabular}
\end{table*}
\begin{table*}[h]
\caption{Precision, Recall, Accuracy, and F-measure for Support Vector Machine, Logistic Regression, Random Forest, and Naive Bayes classifiers for BERT and BioBERT embeddings}
\centering
\begin{tabular}{@{}lllllllll@{}}
\toprule
\textbf{Model} & \multicolumn{4}{c}{\textbf{BERT}} & \multicolumn{4}{c}{\textbf{BioBERT}} \\
\cmidrule(lr){2-5} \cmidrule(lr){6-9}
& \textbf{A} & \textbf{P} & \textbf{R} & \textbf{F} & \textbf{A} & \textbf{P} & \textbf{R} & \textbf{F} \\
\midrule
SVM & 0.9745 & 0.9565 & 0.8995 & 0.9271 & 0.9984 & 0.9851 & 0.9548 & 0.9697 \\
LR & 0.9562 & 0.8745 & 0.9125 & 0.8931 & 0.9628 & 0.9889 & 0.9568 & 0.9726 \\
RF & 0.9235 & 0.9845 & 0.8875 & 0.9335 & 0.9855 & 0.9625 & 0.9789 & 0.9706 \\
NB & 0.932 & 0.895 & 0.9022 & 0.8986 & 0.912 & 0.895 & 0.8625 & 0.8784 \\
\bottomrule
\end{tabular}
\end{table*}
For the TF-IDF feature, SVM has 96.82\%, 94.96\%, 91.28\%, and 92.89\% for the accuracy, precision, recall, and f-measure values, and the LR algorithm has 93.21\%, 98.20\%, 89.77\%, and 93.80\% for the accuracy, precision, recall, and f-measure values, respectively. For the RF classifier, the values were 97.41\%, 95.20\%, 97.45\%, and 96.31\%, and the NB classifier obtained 90.11\%, 97.82\%, 90.01\%, and 93.75\% for the accuracy, precision, recall, and f-measure values. For the CountVector feature, the accuracy, precision, recall, and f-measure values were 92.10\%, 85.96\%, 92.30\%, and 89.02\% for SVM, 89.51.98\%, 85.42\%, 91.02\%, and 88.13\% for LR, 96.58\%, 91.00\%, 89.70\%, and 90.35\% for RF, and 90.11\%, 97.82\%, 90.01\%, and 93.75\% for NB classifier, respectively.\\

\indent For Word2Vec embedding with the SVM classifier, we have obtained 89.45\%, 86.51\%, 91.25\%, and 88.82\% as the accuracy, precision, recall, and f-measure values, and with RF, these values were 97.85\%, 89.70\%, 98.72\%, and 93.99\%, and for NB classifiers, we have obtained 95.65\%, 89.77\%, 95.62\%, and 92.60\%, respectively. For Glove embedding, the SVM has given 79.84\% for accuracy, 89.15\% for precision, 91.20\% for recall, and 90.16\% for F-measure, and the LR algorithm has produced 85.21\% for accuracy, 89.69\% for precision, 91.52\% for recall, and 90.60\% for F-measure. For the RF and NB classifiers, these values were 97.85\%, 94.51\%, 80.47\%, and 86.93\%, and 95.62\%, 91.25\%, 98.85\%, and 94.90\%, respectively. This work also used BERT, a pretrained language model trained on general internet-scale data, for generating the embeddings, and also BioBERT, which was trained on biomedical and healthcare data. For BERT embeddings, SVM has given 97.45\%, 95.65\%, 89.95\%, and 92.71\%, LR has given 95.62\%, 87.45\%, 91.25\%, and 89.31\%, RF has given 92.35\%, 98.45\%, 88.75\%, 93.35\%, and NB has given 93.20\%, 89.50\%, 90.22\%, and 89.86\%, for the precision, recall, accuracy, and f-measure.\\

\indent When the BioBERT embedding was used to train our classifiers, we got a higher f-measure for the SVM, LR, and RF classifiers, but the NB gave only 87.84\% for the f-measure values. This may be due to the assumption of the NB algorithm that the features are independent, but in this sequence understanding task, that is not the case. In conclusion, this work highlighted that a domain-specific transformer-based model will be good in custom named entity recognition tasks, and the same has been showcased for the identification of clinical named entities from the unstructured documents related to Malaria disease. Such an approach may find several interesting applications in the healthcare domain, including automated understanding of clinical narratives and case documents, clinical text summarization, to name a few.
\subsection{Limitations}
The results of this study are promising in the area of named entity recognition for the biomedical domain of Malaria disease, but at the same time, this study has several limitations that should be acknowledged. The corpus used for training and evaluation was limited to malaria-related literature, which may restrict the generalizability of the model to other biomedical domains. The specialized vocabulary and context in malaria research may not fully represent the broader biomedical landscape, potentially affecting the model's transferability. Another limitation is with respect to the contextual embeddings of the BioBERT model, which was trained on large-scale biomedical text. The performance of the model is still dependent on the quality and coverage of the annotated training data, and in this study, the availability of manually annotated malaria-specific NER datasets was limited, which might have affected the ability to capture rare or domain-specific entities. Finally, the model primarily focuses on surface-level textual features without incorporating domain knowledge such as ontologies or structured biomedical databases. This caused challenges in distinguishing between highly similar entities, abbreviations, and synonyms that are frequent in malaria literature.
\section{Conclusions and Future Work}
Biomedical named entity recognition is a problem that has been extensively researched in the area of biomedical NLP, but there is a need for many innovative techniques for the same. In this connection, this work proposed a biomedical named entity recognition approach by fine-tuning BioBERT, a highly developed domain-specific language model specifically designed for the biomedical field, encompassing a vast range of pre-existing knowledge to enhance language understanding and representation within this domain. When the suggested approach's performance was compared to several baselines, the model employing BioBERT embeddings outperformed them for biomedical named entity recognition. As the results are promising, extending the methodology for deep learning approaches with more labeled data seems a promising future work. Also, there are more open-source pretrained biomedical models that are getting published, and using them for generating the embeddings to train the classifiers would be an interesting dimension.
\bibliographystyle{abbrv}
\bibliography{references}

\end{document}